\documentclass[10pt,twocolumn,letterpaper]{article}

\usepackage{cvpr}              
\usepackage{graphicx}
\usepackage{amsmath}
\usepackage{amssymb}

\usepackage[T1]{fontenc}

\usepackage{booktabs}
\usepackage{relsize}
\usepackage{arydshln}
\usepackage[accsupp]{axessibility}  

\definecolor{cvprblue}{rgb}{0.21,0.49,0.74}
\usepackage[pagebackref,breaklinks,colorlinks,allcolors=cvprblue]{hyperref}

\def\paperID{4} 
\def\confName{CVPR}
\def\confYear{2025}

\title{Pose-Aware Weakly-Supervised Action Segmentation}

\author{
  \textbf{Seth Z. Zhao$^1$$^2$}\thanks{These authors contribute equally.} \thanks{Work done as intern at Honda Research Institute, USA.}
  ~~~
  \textbf{Reza Ghoddoosian$^1$}\footnotemark[1] \thanks{Contact: \texttt{reza\_ghoddoosian@honda-ri.com}.}
  ~~~
  \textbf{Isht Dwivedi$^1$} 
  ~~~
  \textbf{Nakul Agarwal$^1$} 
  ~~~
  \textbf{Behzad Dariush$^1$} \\
  $^1$Honda Research Institute, USA~~~~~~$^2$UC Berkeley
}

\begin{document}
\maketitle
\begin{abstract}
Understanding human behavior is an important problem in the pursuit of visual intelligence. A challenge in this endeavor is the extensive and costly effort required to accurately label action segments. To address this issue, we consider learning methods that demand minimal supervision for segmentation of human actions in long instructional videos. Specifically, we introduce a weakly-supervised framework that uniquely incorporates pose knowledge during training while omitting its use during inference, thereby distilling pose knowledge pertinent to each action component. We propose a pose-inspired contrastive loss as a part of the whole weakly-supervised framework which is trained to distinguish action boundaries more effectively. Our approach, validated through extensive experiments on representative datasets, outperforms previous state-of-the-art (SOTA) in segmenting long instructional videos under both online and offline settings. Additionally, we demonstrate the framework's adaptability to various segmentation backbones and pose extractors across different datasets.
\end{abstract}    
\section{Introduction}
\label{sec:intro}
Recognizing human actions in a long instructional videos holds immense significance in facilitating comprehension and learning for intelligent systems. By accurately identifying and understanding human actions depicted in videos, human-machine interaction systems can interpret the sequential steps involved in performing complex tasks. This comprehension aids in skill acquisition and contributes toward enhancing human centered intelligent systems,  human performance evaluation, and monitoring in various industrial applications. One big challenge lies in the fact that frame-level labeling of these videos demands extensive and costly human labor. Thus, a significant amount of research is dedicated to understanding human actions in long videos with  \textit{minimal human-crafted supervision}. 

In this paper we study weakly-supervised learning methods for human action segmentation in long instructional videos. In this context, we work with video sequences paired with an ordered list of action labels (\textbf{transcript}), but without knowing the specific start and end times for each action.
The main focus is to incorporate pose information into a weakly supervised framework, emphasizing its crucial role in temporal segmentation of human actions in the absence of frame-level labels. Pose information is informative due to its ability to encode rich information about body movements, gestures, and interactions, which are essential for a nuanced understanding of human actions. Specifically, pose information enables the decomposition of each action into a series of more granular representations, leading to more discriminative features within the same action segment and aiding in the identification of similarities across different actions. For instance, in an assembly task, the action "fasten screw" can be broken down into reaching for the screw and rotating the screw, each characterized by unique poses despite sharing the same action label. Moreover, it has been shown \cite{pose15} that pose information is particularly discriminative during action transitions, making it a powerful feature to supervise learning and estimate the start and end times of actions in videos with weak labels, where detailed frame-level annotations are absent.

The proposed method leverages both appearance and gesture cues by combining RGB and pose modalities.  While extracting pose information can be beneficial, it also introduces significant computational cost that can hamper real-time performance in interactive applications. To address this issue, we propose a framework that infuses pose information from a standard pose estimator into the RGB frame encoder during training. However, at test time, our method relies exclusively on the RGB modality. In particular, we employ a contrastive learning objective\cite{clip, simclr, simclrv2, supcon, lacon} to enable the model to differentiate between corresponding and non-corresponding RGB and pose features. By establishing a frame-level correspondence between RGB and pose features to create a positive pair, and by introducing a pose-based technique to identify negative pairs, our method facilitates the learning of features within a combined pose-RGB space.

Incorporating pose not only provides supplementary supervision, but also seamlessly distills knowledge into the RGB encoder, eliminating the reliance of pose features during inference. Our experimental evaluations on ATA\cite{Ghoddoosian_2023_ICCV}, IKEA ASM\cite{ben2021ikea}, and Desktop Assembly\cite{desktop} datasets demonstrate versatility and enhanced performance across various segmentation frameworks. The approach remains robust irrespective of the pose extractor used and is effective in both online and offline settings. Here, "online" refers to causal inference in streaming videos for interactive applications, whereas "offline" refers to post-analysis of pre-recorded videos.

The contributions of this paper are summarized as follows:
\begin{itemize}
\item This work is the first to integrate pose information into a weakly-supervised action segmentation framework establishing a segmentation approach, where despite the reliance on pose data during training, the model performs inference using only the conventional RGB modality, as commonly referenced in the literature~\cite{viterbi,CDFL,tasl,Ghoddoosian_2023_ICCV}.
\item We introduce our pose-based contrastive learning loss to distill pose knowledge into the RGB encoder, enhancing its capability to detect action boundaries in weakly-labeled untrimmed videos. This is achieved by utilizing the raw pose similarity across different frames to identify  negative pairs for our loss.
\item Through comprehensive ablation studies and rigorous testing on a variety of video datasets, including ATA, IKEA ASM, and Desktop Assembly, we demonstrate the versatility and broad applicability of our approach. Our method not only results in performance improvement across different segmentation frameworks and pose extraction tools but also proves effective in both online and offline scenarios. 
\end{itemize}

\section{Related Works}
\label{sec:relatedwork}

\subsection{Weakly-Supervised Action Segmentation}
Training action segmentation models under the weak supervision of transcripts was mainly initiated by \cite{bojanowski2014weakly}. Since then, many others \cite{reza:HM,kuehne2017weakly,Fine2Coarse,viterbi,chang2021dpdtw} have proposed iterative \cite{Fine2Coarse,TCFPN,viterbi,CDFL,tasl,Ghoddoosian_2023_ICCV} or end-to-end \cite{d3tw,mucon2021} approaches to align video frames to a given sequence of actions during training. However, they only use RGB-based features (I3D\footnote{For simplicity, without loss of generality, we consider I3D features RGB-based although more accurately they are a mix of RGB and optical flow\cite{OF} streams.}\cite{I3D}  or iDT\cite{iDT}) as input during both ``inference and training''. More similar to us, \cite{Ghoddoosian_2022_CVPR} use multi camera view points, only in training, to estimate more accurate frame-level pseudo labels. Consequently, they can segment videos using single view point input at test time. In contrast to all previous methods, we are the first to utilize pose to guide training and instill skeleton knowledge to the standard RGB-based features for inference.

\subsection{Pose in Action Understanding}

There has been extended research in exploiting pose information for various video understanding tasks. Many papers focus on skeleton-based action recognition \cite{pose2,pose3,pose4,pose5,pose6,pose7,pose8,pose9}, detection \cite{pose10,pose12,pose18}, and anomaly detection \cite{pose12}. In these works pose information is used as the sole input \cite{pose2,pose3,pose4,pose5,pose6,pose7,pose10,pose12} or combined with RGB frames \cite{pose8,pose9,pose18} to classify actions. In addition, \cite{pose11} and \cite{pose13} have further utilized contrastive loss between text and pose representations for action recognition and anomaly detection, respectively, in short videos. Similar to us, \cite{pose14} apply pose to action segmentation. Specifically, they improve a skeleton estimator using self-supervised generative models. However, unlike our framework, all the aforementioned methods use pose in both training and inference time. 

Our work is more aligned with recognition and detection methods that distill cross-modal knowledge from optical flow \cite{pose16,pose22}, pose \cite{pose15,pose19,pose23,reilly2024just} or depth \cite{pose20,pose21} to RGB encoders during training, so that at test time no modality except RGB is required. However, these methods are trained on fully labeled videos. In particular,  although \cite{pose17} does not require annotations for distillation through a pose reconstruction loss, they assume that an approximate bounding box for the athlete is provided in each frame. Also, fully-labeled sports videos are used for training in \cite{pose17} for action recognition. Meanwhile, we are the first to take advantage of pose to guide weakly-supervised training and segment long videos into fine grained actions. 
In the area of self supervised learning, \cite{STEPs} uses cross modal similarity/dissimilarity such that features corresponding to all frames of a segment lie close in the latent space. In contrast, our RGB encoder learns to breakdown such segments into more discriminative pose-based representations by distilling pose knowledge in training.

\subsection{Contrastive Learning in Video Understanding}
Contrastive learning is a popular solution for learning strong representations among multimodal interactions in both pretraining and multi-tasking settings \cite{yang2022vision, ALBEF, blip, pevl, xu2022pretram, SMD, clip, cl_speech}. At a high level, contrastive learning contrasts samples against each other to learn features that are common and different between labels. Introduced in \cite{clip}, CLIP is pretrained using a constrastive loss function to learn image representations from text. \cite{supcon} expands on the contrastive objective of CLIP by producing challenging negative captions for every image-caption pair and selecting robust alternative images. \cite{lacon} proposed a triplet contrastive loss objective based on InfoNCE \cite{simclr} to draw together the embeddings of corresponding image-text pairs, while simultaneously separating non-matching pairs. In the video understanding domains, contrastive learning is often used as a pretraining objective in \cite{xu-etal-2021-videoclip, xu-etal-2021-vlm, Castro_2022_BMVC, movieclip, lin2022frozen}. These methods often facilitate the representation learning of videos by leveraging vast amount of language transcripts \cite{xu-etal-2021-videoclip, xu-etal-2021-vlm} or simply transferring the knowledge learned from image-text alignments \cite{lin2022frozen}. In our work, we aim to leverage contrastive learning methods to facilitate video representation learning with the help of pose features.
\begin{figure*}[t]
\centering
 \includegraphics[width=0.95\textwidth,keepaspectratio]{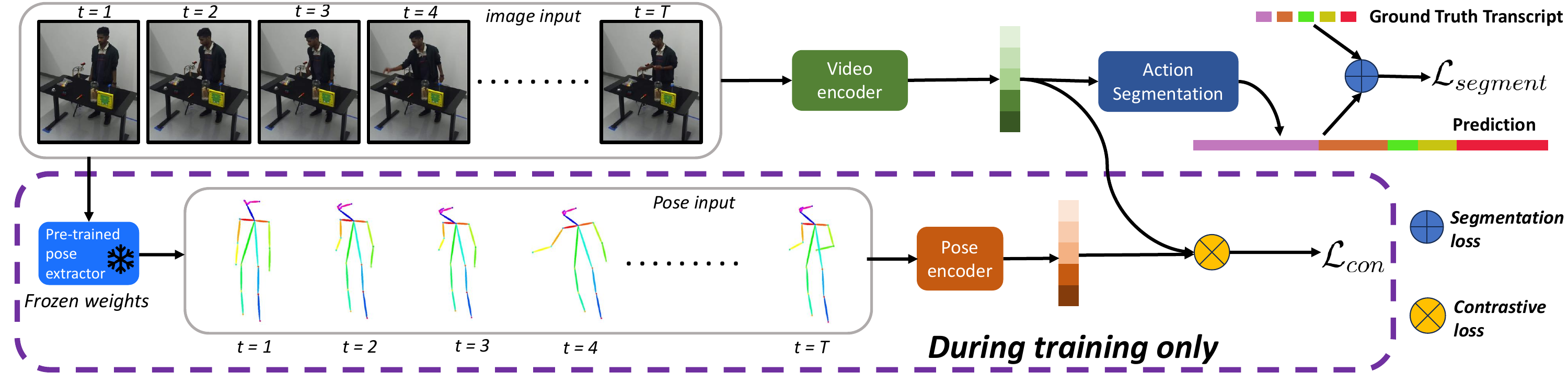}
\caption{Framework overview: Pose information is used exclusively during training. During inference, only image input is considered, omitting the pose branch.}
\label{fig:framework}
\end{figure*}
\section{Method}
\label{sec:method}
In this section, we present the problem formulation and an overview of the proposed pipeline. We then detail the pose encoding process and elaborate on the proposed contrastive losses.  

\subsection{Problem Formulation}
We formulate our task of weakly-supervised video action segmentation as follows. Given a video $\mathbf{x}_1^{t} = (x_1, ..., x_t)$ with $t$ frames and a ``single person'', the goal of the segmentation model is to segment a test video into a sequence of $n$ actions $\mathbf{a}_1^n = (a_1, ..., a_n)$ and their duration $\mathbf{l}_1^n = (l_1, ..., l_n)$. Notice that in a weakly-supervised training setting, we are not given frame-level action labels and we could only assume a sequence of action labels (transcripts) $\mathcal{T}_1^{n} = (\mathcal{T}_1, ..., \mathcal{T}_n)$ that occur throughout the video. During the inference stage, two modes of offline and online settings are employed following previous protocols \cite{viterbi, Ghoddoosian_2022_CVPR, Ghoddoosian_2023_ICCV}. In offline mode, the model processes the entire video before segmentation, whereas in online mode, the model segments in real-time, only accessing frames up to the current moment.

\subsection{Method Overview}
Our approach leverages pose features for enhanced supervision, improving visual representations learning without requiring per-frame action labels. As depicted in Fig. \ref{fig:framework}, we input precomputed RGB features and human poses, extracted by any frozen off-the-shelf estimator, into our framework. These inputs are processed by individual shallow encoders and then mapped into a shared representation space with consistent feature dimensions. Subsequently, contrastive learning loss is applied to embeddings from both modalities, enabling the RGB encoder to learn semantically rich visual representations enriched by pose data during training.
This RGB encoder is also shared with a chosen weakly-supervised segmentation framework to decode the final output. 
During training, in order to integrate our contrastive learning with the original segmentation task, we use a multi-task setting. Here, the model minimizes a joint optimization objective, allowing the RGB encoder to incorporate the pose data and steer the segmentation loss to identify correct action segments across the video. The resulting training loss is:
\begin{equation}
\label{training_loss}
\mathcal{L}_{Final} = \mathcal{L}_{con} + \mathcal{L}_{segment}, 
\end{equation}
where $\mathcal{L}_{con}$ is our proposed pose-based contrastive loss and $\mathcal{L}_{segment}$ is the segmentation loss adopted from any weakly-supervised segmentation baseline \cite{viterbi, Ghoddoosian_2022_CVPR, Ghoddoosian_2023_ICCV}.
During inference, we solely employ the RGB encoder, omitting the pose stream entirely and making our pipeline generalizable to various baselines without impacting runtime performance.

\subsection{Detailed Pipeline}
In this section, the process of extracting pose embeddings is first explained, followed by how pose information is infused into the RGB embeddings through our contrastive learning method in untrimmed videos.
\subsubsection{Pose Encoding}
Given a frame at time $t$, raw pose $p_t \in \mathbb{Z}^{K\times 2}$ is a collection of $(x,y)$ coordinates for $K$ human keypoints. Here, $K$ represents the number of 2D keypoints extracted by an external pose extractor and $\mathbb{Z}$ is the set of integers. Before inputting these raw keypoints to the pose encoder, we perform a normalization step to ensure they are unaffected by changes in perspective, rotation, and positional offset in the frame. Specifically, each keypoint is centered and scaled with respect to the "center of mass" of the human, which is determined by averaging the coordinates of all joints. Subsequently, we determine the angle required to rotate each adjusted keypoint so that the head and "center of mass" align vertically, sharing the same $x$ coordinates.  These normalized 2D keypoints, $\overline{p}_t$, are then fed into the pose encoder. 

As shown in Eqs. \ref{eq:pose_ecoder1}-\ref{eq:pose_ecoder3}, the encoder uses a light-weight two-layer MLP network to learn rich representations from the pose keypoints and map them to the joint RGB-pose space. Following the approach in~\cite{pose11}, each encoder layer is structured with sequential steps of layer normalization, ReLU activation, and dropout, with a residual link between layers complemented by max-pooling and a linear projection function $\Gamma$ to refine the dimensionality of the resultant pose embedding $P_t$. Further details of the pose normalization and encoder architecture can be found in the supplementary materials.

\begin{align}\label{eq:pose_ecoder1}
z_1 &= \text{dropout}(\text{ReLU}(\text{LayerNorm}(W_1\overline{p}_t + b_1))),  \\
z_2 &= \text{dropout}(\text{ReLU}(\text{LayerNorm}(W_2z_1 + b_2))), \\
P_t &= \Gamma\big(\text{maxpool}(z_2 + z_1)\big).
\label{eq:pose_ecoder3}
\end{align}

\subsubsection{Pose-Supervised Contrastive Learning}

In this paper, we aim to utilize contrastive loss to create a common space for embedding both pose and RGB data. This guides the RGB encoder to learn and infuse pose information into its output embeddings from processing RGB frames alone. Given a video $\mathbf{V}_1^{T} = (v_1, ..., v_T)$ with $T$ frames, we extract frame-level pose embeddings $\mathbf{P}_1^{T} = (P_1, ..., P_T)$, and RGB embeddings  $\mathbf{I}_1^{T} = (I_1, ..., I_T)$, where $I_t$ is the output of the RGB encoder at frame $t$. Various networks, such as Transformers or CNNs, can be utilized to implement the RGB encoder. For each frame $t$, serving as the anchor frame in one modality (RGB or pose), we define $\mathbb{A}(t)$ and $\overline{\mathbb{A}}(t)$ as the sets of positive and negative frames, respectively, from the other modality within the same video. These sets, $\mathbb{A}(t)$ and $\overline{\mathbb{A}}(t)$, help identify positive and negative instances from the alternate modality that form corresponding pairs with the anchor at frame $t$. Utilizing these pairings for the anchor frame $t$, the RGB to pose contrastive loss $\mathcal{L}_{I2P}$ is designed to increase similarity in positive pairs and dissimilarity in negative pairs.

\begin{small}
\begin{equation}
\label{contrastive_loss_I2P}
 \mathcal{L}_{I2P} = - \frac{1}{T} \mathlarger{\sum}_{t \in [0, T)} \log \frac{\sum_{i \in \mathbb{A}(t)} exp(sim(I_t, P_i)/\tau)}      { \sum_{j \in \{\mathbb{A}(t) \cup \overline{\mathbb{A}}(t)\}} exp(sim(I_t, P_j)/\tau)},
\end{equation}
\end{small}

where $\tau$ is the temperature parameter and $sim$ denotes the similarity function. Similarly, we derive the pose to RGB contrastive loss $\mathcal{L}_{P2I}$:
\begin{small}
\begin{equation}
\label{contrastive_loss_P2I}
\mathcal{L}_{P2I} = - \frac{1}{T} \mathlarger{\sum}_{t \in [0, T)} \log \frac{\sum_{i \in \mathbb{A}(t)} exp(sim(P_t, I_i)/\tau)}      {\sum_{j \in \{\mathbb{A}(t) \cup \overline{\mathbb{A}}(t)\}} exp(sim(P_t, I_j)/\tau)}.
\end{equation}
\end{small}

Our overall contrastive loss $\mathcal{L}_{con}$ is defined as the sum of $\mathcal{L}_{I2P}$ and $\mathcal{L}_{P2I}$. 
Since each video encompasses various actions without frame-level labels, identifying the set of positive and negative frames, $\mathbb{A}(t)$ and $\overline{\mathbb{A}}(t)$, for the contrastive loss $\mathcal{L}_{con}$ poses a challenge.

Given anchor frame at $t$, a vanilla method is matching in time across different modalities to create a positive pair, and consider any other frame from the other modality as a negative pair. Formally, $\mathbb{A}(t)=\{t\}$ and $\overline{\mathbb{A}}(t)=\{j | j\in [0,T) \wedge j\neq t \}$.

\begin{figure}[t]
\centering
 \includegraphics[width=0.44\textwidth,keepaspectratio]{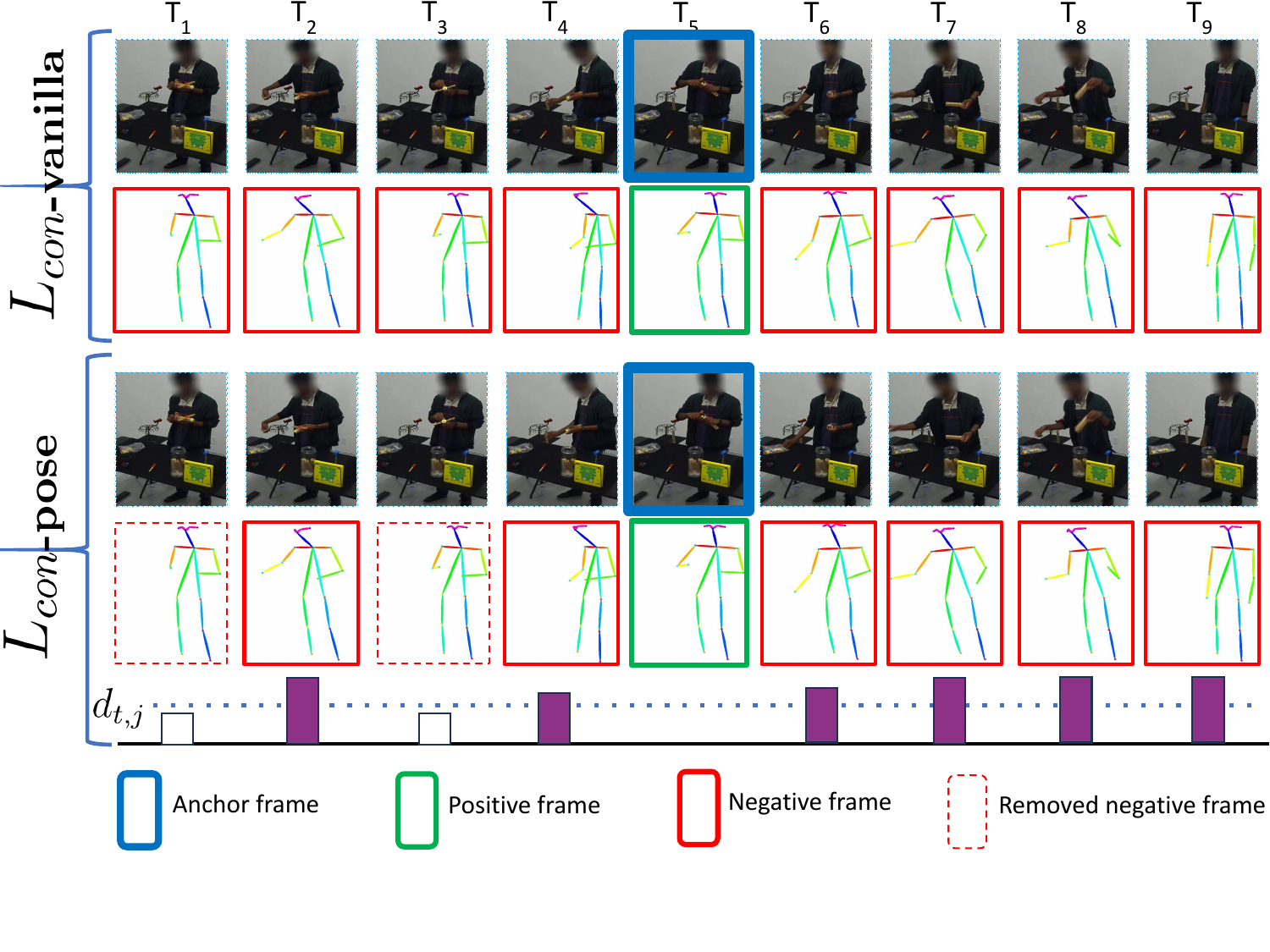}
\caption{Two methods to mine positive and negative frames for a given anchor in our contrastive learning framework.}
\label{fig:contrastive}
\end{figure}

While this vanilla contrastive learning is straightforward, it suffers from identifying false negative pairs under two primary conditions. Firstly, in instances where a pose is held for an extended period within a specific action sequence. For example, consider the RGB embedding at frame $t$ during the action ``balance part'' as the anchor. Due to the static nature of this action, many negative poses identified throughout this segment closely resemble the positive pose at time $t$. The second challenging scenario occurs when a similar pose is repeated across various actions and is incorrectly tagged as a negative pose simply because it appears at a different time than $t$.  In the context of pose-supervised contrastive learning, we maintain that the infused pose knowledge should not depend on the action labels,  especially in the absence of ground-truth. Since pose configurations are building blocks of actions and can reoccur across different actions, it is important to transfer the pose knowledge into the RGB encoder without linking it to specific action categories. Consequently, we filter out any negative pairs that feature poses similar to the anchor's, regardless of their occurrence time (see Fig. \ref{fig:contrastive}). To achieve this, we introduce $d_{t,j}=|\overline{p}_t-\overline{p}_j|$ as the distance between the normalized key points of frames $t$ and $j$. Accordingly, we redefine the set of negative frames as $\overline{\mathbb{A}}(t)=\{j | j\in [0,T) \wedge d_{t,j}\geqq \delta   \}$, where $\delta $ is a predefined threshold. $\mathbb{A}(t)=\{t\}$ is the same as in the vanilla case. 

\section{Experiments}
\label{sec:experiments}

In this section, we describe our experiments, followed by ablation studies and results of our framework compared to various baselines on multiple datasets. In the end, we conclude with qualitative evidence of how pose contributes to more accurate temporal boundary detection. Additional experiments are included in the supplementary material.
\subsection{Experimental Settings}
\noindent \textbf{Datasets.} We conduct our experiments on three publicly available instructional video datasets: the ATA dataset\cite{Ghoddoosian_2023_ICCV}, the Desktop Assembly dataset\cite{desktop}, and the IKEA dataset\cite{ben2021ikea}. The ATA dataset contains 1152 toy assembly videos, captured from four different view-points, with each video averaging 1.3 minutes long and 12.9 segments. It features 32 participants assembling three different toys with 15 action classes and 96 unique transcripts. We adhere to the standard subject-based splitting of this dataset for test and validation. The Desktop Assembly dataset contains 76 desktop assembly videos, amounting a total of 2 hours, annotated with 23 action classes and 6 similar transcripts. It is split into 59 training and 17 testing videos. Lastly, the IKEA dataset contains 1113 furniture assembly videos, recorded from three perspectives, with an average duration of 1.9 minutes. This dataset is categorized into 33 action classes and offers 5 different training/testing splits. 

\noindent \textbf{Evaluation Metrics.} Following previous work~\cite{3rec, Ghoddoosian_2023_ICCV, Ghoddoosian_2022_CVPR}, we use four metrics to evaluate our action segmentation performance.  1) \textit{acc} represents the average frame-level accuracy. 2) \textit{IoU} determines intersection-over-union ratio for each predicted segment, excluding the background frames. 3) \textit{Edit} employs edit distance to assess the similarity between predicted and ground-truth transcripts. 4) \textit{F1@0.5} assesses the per-class F1 score for predicted segments with an IoU threshold of 0.5. 

\noindent \textbf{Implementation Details.} To remain consistency with prior works, we extracted I3D \cite{Carreira2017QuoVA} features from ATA and IKEA datasets, and for Desktop Assembly, we used ResNet \cite{ResNet} features. In experiments with DP\cite{Ghoddoosian_2023_ICCV} as the baseline, we modeled the video encoder with Transformers. For MuCon\cite{mucon2021} and TASL\cite{tasl} segmentation baselines, we used their existing temporal convolution and GRU network outputs for RGB embedding, respectively. For computational efficiency, pose keypoints were extracted every five frames by RTMPose Body2D \cite{rtmpose}. $\delta=0.15$ for the experiments on the Desktop dataset while for ATA, $\delta$ is set to 0.05 and 0.2 for online and offline segmentation respectively. The effect of $\delta$ is discussed in Section \ref{ablation}. All other parameters, such as the number of training iterations, are set as per baseline settings~\cite{Ghoddoosian_2023_ICCV,mucon2021,tasl}. More implementation details are included in the supplementary material. We intend to release the code and all parameters upon acceptance.

\subsection{Comparison Results}
In this section, we show our pose-supervised segmentation framework improves both online and offline results on multiple datasets and baselines.  

\begin{table}[t]
    \caption{Main results on weakly-supervised online segmentation. Our proposed pose-inspire framework improves the previous methods across different datasets.}
    \centering
    \resizebox{0.45 \textwidth}{!}{
    \begin{tabular}{l|l|c|c|c|c}
    \toprule
        Dataset & Method & \textit{acc} & \textit{IoU} & \textit{Edit} & \textit{F1@0.5} \\ 
    \midrule
        \text{ATA}\cite{Ghoddoosian_2023_ICCV} 
        & \text{Greedy}\cite{gao2021woad} & 60.2 & 53.5 & 47.8 & 41.4 \\
        & \text{DP}\cite{Ghoddoosian_2023_ICCV} &  62.3 & 53.3 & 55.5 & 48.2 \\
        & \text{DP + Ours} &  \textbf{66.0}  & \textbf{58.7} & \textbf{56.9} & \textbf{51.2} \\
    \midrule
        \text{Desktop}\cite{desktop} 
        & \text{Greedy} \cite{gao2021woad} & 4.8 & 2.5 & 24.7 & 0.3 \\
        & \text{DP}\cite{Ghoddoosian_2023_ICCV} & 10.5 & 5.1 & 36.8 & 2.3 \\
        & \text{DP + Ours} &  \textbf{18.0}  & \textbf{7.6}  & \textbf{52.2} & \textbf{3.7} \\
    \midrule
        \text{IKEA}\cite{ben2021ikea} 
        & \text{Greedy} \cite{gao2021woad} & 53.0 & 27.0 & 41.5 & 23.6 \\
        & \text{DP}\cite{Ghoddoosian_2023_ICCV} &  54.3 & 27.3 & 48.1 & 26.0 \\
        & \text{DP + Ours} &  \textbf{54.4}  & \textbf{27.7} & \textbf{48.4} & \textbf{26.2} \\
    \bottomrule
    \end{tabular}}
    \label{tab:main_results_online}
\end{table}

\noindent \textbf{Weakly-Supervised Online Segmentation.} In Table \ref{tab:main_results_online}, we demonstrate the impact of our pose-inspired contrastive loss in comparison with previous weakly-supervised online segmentation methods. Notice that during training, both Greedy and DP share the same network structure. However, at inference time, Greedy adopts a sliding window approach to predict per-frame actions while DP uses an unconstrained dynamic programming approach based on the available transcripts. As shown in Table \ref{tab:main_results_online}, infusing pose information into the RGB encoder of DP elevates its  performance across all four metrics and three datasets. Specifically, on ATA  and Desktop Assembly Dataset, the \textit{IoU} performance gain is about 5.7\% and 2.1\%, respectively. We associate the smaller improvements on the IKEA dataset mostly to its 5th split. In many videos of this split, the single person assumption is violated by background people, which negatively impacts our pose encoding accuracy. We provide split-wise results on the IKEA dataset in the supplementary material.

\begin{table}[t]
    \caption{Main results on weakly-supervised offline segmentation. Our pose-inspired framework is integrated into different baselines, which results in performance improvement across different metrics and datasets. Results are based on the best $\delta$ values.}
    \centering
        \renewcommand{\arraystretch}{1.07} 

    \resizebox{0.45 \textwidth}{!}{
    \begin{tabular}{l|l|c|c|c|c}
    \toprule
        Dataset & Method & \textit{acc} & \textit{IoU} & \textit{Edit} & \textit{F1@0.5} \\ 
    \midrule
        \text{ATA}\cite{Ghoddoosian_2023_ICCV} 
        & \text{CDFL} \cite{CDFL} & 58.1 & 44.9 & 59.5 & 50.9 \\
        \cdashline{2-6}
        & \text{MuCon} \cite{mucon2021} & 46.4 & \textbf{33.5} & 53.7 & 32.2 \\
        & \text{MuCon + Ours} & \textbf{48.3} & 32.3 & \textbf{54.5} & \textbf{33.5}\\
        \cdashline{2-6}
        & \text{TASL} \cite{tasl} & 39.3 & 27.5 & \textbf{55.7} & 27.5 \\
        & \text{TASL + Ours} & \textbf{45.7} & \textbf{29.3} & 51.1 & \textbf{33.2} \\
        \cdashline{2-6}
        & \text{DP} \cite{Ghoddoosian_2023_ICCV} & 65.1 & 55.7 & 65.5 & 59.3 \\
        & \text{DP + Ours} & \textbf{68.5} & \textbf{61.0} & \textbf{69.5} & \textbf{63.8} \\
    \midrule
        \text{Desktop}\cite{desktop}
        & \text{CDFL} \cite{CDFL} & 16.5 & 10.7 & 81.9 & 7.2 \\
        \cdashline{2-6}
        & \text{MuCon} \cite{mucon2021} & 46.0 & 33.0 & 100.0 & 27.2 \\
        & \text{MuCon + Ours} & \textbf{50.9} & \textbf{35.4} & \textbf{100.0} & \textbf{31.3} \\ 
        \cdashline{2-6}
        & \text{TASL} \cite{tasl} & 35.2 & 22.4 & 95.8 & 14.7 \\
        & \text{TASL + Ours} & \textbf{41.2} & \textbf{27.1} & \textbf{96.6} & \textbf{20.7} \\
        \cdashline{2-6}
        & \text{DP} \cite{Ghoddoosian_2023_ICCV} & 16.3 & 10.7 & 86.5 & 6.5 \\
        & \text{DP + Ours} & \textbf{17.2} & \textbf{12.0} & \textbf{91.4} & \textbf{7.7} \\
    \bottomrule
    \end{tabular}}
    \label{tab:main_results_offline}
\end{table}

\noindent \textbf{Weakly-Supervised Offline Segmentation.} For the sake of completeness, we also integrate our pose-inspired contrastive framework into three state-of-the-art offline segmentation methods, i.e., DP\cite{Ghoddoosian_2023_ICCV}, TASL\cite{tasl}, and MuCon\cite{mucon2021}. As shown in Table \ref{tab:main_results_offline}, infusing pose can consistently improve their weakly-supervised performance on the ATA and Desktop Assembly datasets.  Despite the differences in network architecture and segmentation technique among various methods, our framework can be adapted to all baselines without changing their original architecture. In particular, on Desktop Assembly videos, instilling pose knowledge into the MuCon encoder achieves new SOTA and improves \textit{acc} and \textit{F1} by up to approximately 5\%. Also, the high \textit{Edit} score on Desktop Assembly videos is due to the very similar 6 transcripts of this dataset. Conversely, DP stands out as the best baseline for ATA videos, owing to its design for segmenting unseen sequences in the ATA test set. 

\subsection{Analysis and Ablation Studies}
\label{ablation}
In this section, we first compare the performance of our proposed contrastive loss, then test our framework's robustness across various pose extractors, and finally examine the pose knowledge learned by the RGB encoder. We use DP~\cite{Ghoddoosian_2023_ICCV} and TASL\cite{tasl} as baselines for our ablation study in online and offline segmentation tasks respectively.

\begin{table}[t]
    \caption{Comparison of vanilla and pose-inspired contrastive learning in weakly-supervised segmentation in the Desktop dataset. Results with the DP and TASL baselines correspond to online and offline segmentation modes respectively.}
    \centering
    \resizebox{0.45 \textwidth}{!}{
    \begin{tabular}{l|l|c|c|c|c}
    \toprule
         Backbone & Method & \textit{acc} & \textit{IoU} & \textit{Edit} & \textit{F1@0.5} \\
    \midrule
          & Baseline & 10.5 & 5.1 & 36.8 & 2.3 \\
        \text{DP} \cite{Ghoddoosian_2023_ICCV}& $L_{con\text{-vanilla}}$ & 12.7  & 7.2  & 48.3 & 3.7 \\
         & $L_{con\text{-pose}}$ & \textbf{18.0} & \textbf{7.6} & \textbf{52.2} & \textbf{3.7}  \\
        \cline{1-6}
          & Baseline & 35.2 & 22.4 & 95.8 & 14.7 \\
        \text{TASL} \cite{tasl}& $L_{con\text{-vanilla}}$
        & 39.3 & 26.3 & 95.8 & 16.5 \\
         & $L_{con\text{-pose}}$ & \textbf{41.2} & \textbf{27.1} & \textbf{96.6} & \textbf{20.7} \\
    \bottomrule
    \end{tabular}}
    \label{tab:clip_vs_supcon_vs_threshold}
\end{table}

\noindent \textbf{Contrastive Learning Mining Techniques.} We compare the results of our proposed pose-based and vanilla mining techniques in Table \ref{tab:clip_vs_supcon_vs_threshold} for both online and offline segmentation. In particular, we utilized DP~\cite{Ghoddoosian_2023_ICCV} for online and TASL\cite{tasl} for offline segmentation tasks. Both contrastive learning methods outperform the baseline in all weakly-supervised segmentation experiments, demonstrating how the RGB encoder significantly benefits from the pose knowledge infusion. In addition, Table \ref{tab:clip_vs_supcon_vs_threshold} shows that vanilla learning is consistently inferior to the pose-supervised method, as it introduces a higher number of false negative samples that confuse the RGB model. On the other
hand, utilizing pose for mining negative and positive instances account for pose variations within the same segment. This leads to a more fine-grained understanding of human dynamics and improves the recognition of the class and time extent of each segment in long videos.

\begin{figure*}[t]
        \centering
\includegraphics[width=0.9\linewidth]{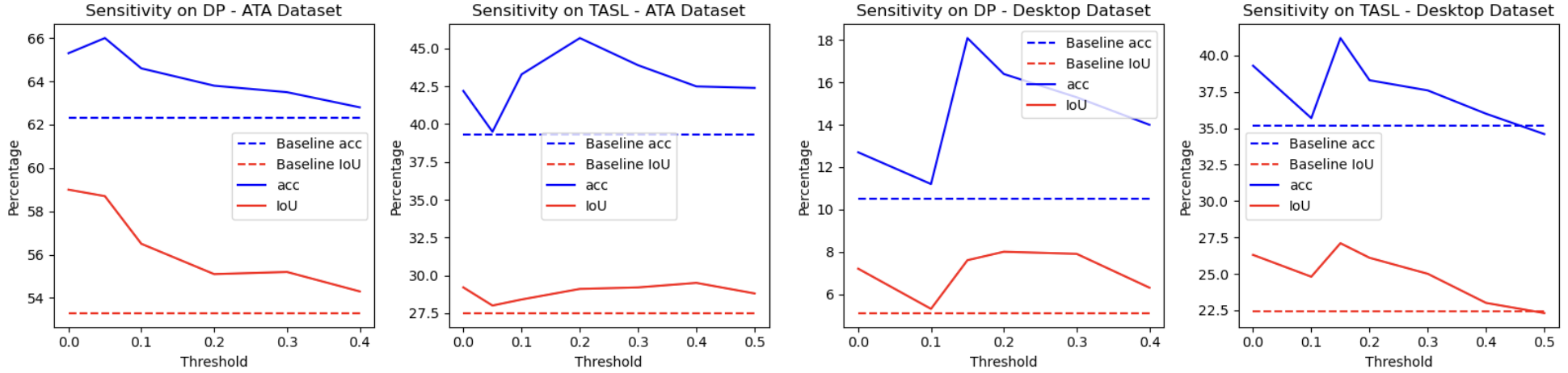}
    \caption{Sensitivity analysis of $L_{con\text{-pose}}$ on DP\cite{Ghoddoosian_2023_ICCV} online segmentation and TASL\cite{tasl} offline segmentation on both ATA\cite{Ghoddoosian_2023_ICCV} and Desktop\cite{desktop} datasets. Note that x-axis represents threshold value and y-axis represents results of \textit{acc} and \textit{IoU}.} 
\label{fig:sensitivity}
\end{figure*}

The sensitivity of the threshold $\delta$ in the pose-supervised learning method is illustrated in Fig. \ref{fig:sensitivity}. Notably, incorporating the pose-supervised loss enhances performance over the baseline across all threshold values. $\delta$ varies from 0, where no negative frames are removed, to a sufficiently high value that leads to the removal of all negative samples for contrastive learning. As shown in Fig. \ref{fig:sensitivity}, results converge to the baseline as all negative samples are removed.  Also, Note that the vanilla approach is a special case of the pose-supervised method when $\delta=0$. 

The statistics of the pose distance $d_{t,j}$ between any two frames $t$ and $j$ of a video is sensitive to the view-point. Hence, in a dataset like ATA, which features multiple viewpoints, finding a fixed effective threshold across all views is challenging. This is because a threshold value that is low for one view may be too high for another, leading to the removal of true negative frames.

\begin{figure*}[t]
        \centering
\includegraphics[width=0.9\linewidth]{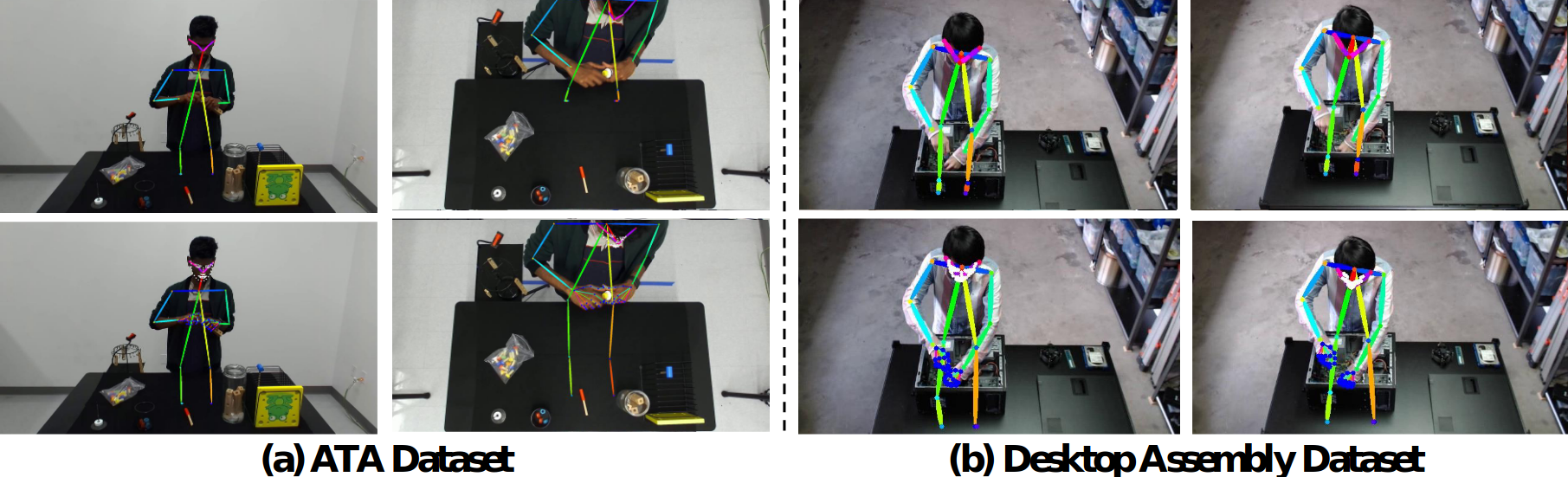}
    \caption{Visualization of zero-shot pose extraction results on both ATA Dataset and Desktop Assembly Dataset. Note that first row and second row represent RTMPose Body2D (sparse keypoints) and RTMPose WholeBody2D (dense keypoints) results, respectively. Compared to the Body2D keypoints, Wholebody2D keypoints have 116 additional keypoints on hands and face.} 
\label{fig:sparse_and_dense}
\end{figure*}

\noindent \textbf{Pose Type Generalizability.} We assess the robustness of our framework by examining its online and offline performances with three pose extractors. We employ RTMOPose extractor\cite{lu2023rtmo}, RTMPose Body2D, and RTMPose WholeBody2D extractors to integrate pose into our framework. The main difference between these pose extractors is the level of keypoint detail, as illustrated in Fig. \ref{fig:sparse_and_dense}. The RTMPose WholeBody2D extractor identifies 133 fine-grained keypoints across the face, hands, and body, whereas RTMPose Body2D and RTMOPose identify 17 sparser set of keypoints. As shown in Table \ref{tab:various_pose_table}, our framework outperforms DP and TASL baselines on both the ATA and Desktop Assembly datasets, regardless of the extractor used, indicating its adaptability to different levels of pose detail. Table \ref{tab:various_pose_table} further suggests that a higher number of keypoints results in competitive or larger improvements, due to the more detailed pose representations. This improvement is more substantial in the Desktop Assembly dataset where fine-grained pose estimations are more accurate. Also, note that while RTMOPose\cite{lu2023rtmo} achieves faster inference speed, its performance on open-world human pose extraction is not as accurate as RTMPose\cite{rtmpose} series.

\begin{table}[tb]
    \caption{Results of our pose-supervised framework using RTMPose Body2D\cite{rtmpose}, RTMPose WholeBody2D\cite{rtmpose}, and RTMOPose\cite{lu2023rtmo} pose detectors. Note that for DP\cite{Ghoddoosian_2023_ICCV} and TASL\cite{tasl} we use online and offline segmentation settings respectively. Our framework improves the performance compared to the baselines \cite{Ghoddoosian_2023_ICCV, tasl} irrespective of the external pose extractor.}
    \centering
    \resizebox{.45\textwidth}{!}{
    \begin{tabular}{l|l|l|c|c|c}
    \toprule
        Dataset & Pose Extractor & \textit{acc} & \textit{IoU} & \textit{Edit} & \textit{F1@0.5} \\
    \midrule
        \text{ATA}\cite{Ghoddoosian_2023_ICCV}
        & DP \cite{Ghoddoosian_2023_ICCV} (No Pose) &  62.3 & 53.3 & 55.5 & 48.2 \\
        & with \text{RTMPose Body2D} &  \textbf{66.0}  & \textbf{58.7} & \textbf{56.9} & \textbf{51.2}  \\
        & with \text{RTMPose WholeBody2D} &  64.6  & 57.5 & 55.8 & 50.6 \\
        & with \text{RTMOPose} & 63.2 & 56.2 & 55.9 & 48.6 \\
    \midrule
        \text{Desktop}\cite{desktop}
        & DP \cite{Ghoddoosian_2023_ICCV} (No Pose) &  10.5 & 5.1 & 36.8 & 2.3 \\
        & with \text{RTMPose Body2D} &  18.0  & 7.6  & 52.2 & 3.7 \\
        & with \text{RTMPose WholeBody2D} &  \textbf{22.8} & \textbf{14.1} & \textbf{53.9} & \textbf{10.8} \\
        & with \text{RTMOPose} & 12.1 & 7.0 & 39.2 & 3.1 \\
 \specialrule{.2em}{.1em}{.1em}
        \text{ATA}\cite{Ghoddoosian_2023_ICCV} 
        & TASL \cite{tasl} (No Pose) &  39.3 & 27.5 & \textbf{55.7} & 27.5 \\
        & with \text{RTMPose Body2D} &  \textbf{45.7}  & \textbf{29.3} & 51.1 & 33.2  \\
        & with \text{RTMPose WholeBody2D} & 44.9 & 28.2 & 55.1 & \textbf{34.5} \\
        & with \text{RTMOPose} & 39.5 & 28.2 & 49.9 & 32.0 \\
    \midrule
        \text{Desktop}\cite{desktop}
        & TASL \cite{tasl} (No Pose) &  35.2 & 22.4 & 95.8 & 14.7 \\
        & with \text{RTMPose Body2D} &  41.2  & 27.1  & \textbf{96.6} & 20.7 \\
        & with \text{RTMPose WholeBody2D} & \textbf{41.6} & \textbf{28.0} & 95.8 & \textbf{23.3} \\
        & with \text{RTMOPose} & 36.3 & 24.6 & 96.0 & 18.7 \\
    \bottomrule
    \end{tabular}}
    \label{tab:various_pose_table}
\end{table}

\noindent \textbf{Pose Knowledge Transferability.}
In this section, we explore the extent to which pose knowledge, acquired during training, is applied during inference in the absence of an explicit pose modality. To investigate this, we conduct an experiment where, rather than infusing pose knowledge, we extract pose keypoints during both training and inference phases. 
In this setup, pose and RGB embeddings are merged prior to input into the segmentation model, and trained with the same loss as our proposed method to allow for a direct comparison. The concatenation baseline serves as the upper bound of our proposed method. Table \ref{tab:fusion_strategy_table} shows the RGB encoder in our method effectively assimilates pose knowledge through contrastive learning, often yielding performance comparable to its upper bound, even without direct use of pose information during inference.

Additionally for more insight, in Table \ref{tab:only_pose_table}, we compare our pose to RGB distillation result to that of RGB to pose distillation as well as pose and RGB only segmentation results. Table \ref{tab:only_pose_table} shows that  pose features are less discriminative than RGB features for action segmentation. While the performance of pose-only inference (row 1) is improved upon RGB distillation (row 2), it can not still compete with even the RGB-alone baseline (row 3). 

\begin{table}[tb]
    \caption{Comparison of pose knowledge distillation (no pose at inference) with the oracle (\text{DP} + Concat) where pose is used at inference too on weakly-supervised online segmentation. }
    \centering
    \resizebox{.47\textwidth}{!}{
    \begin{tabular}{l|l|c|c|c|c}
    \toprule
        Dataset & Method & \textit{acc} & \textit{IoU} & \textit{Edit} & \textit{F1@0.5} \\
    \midrule
        \text{ATA}\cite{Ghoddoosian_2023_ICCV} 
        & \text{DP}\cite{Ghoddoosian_2023_ICCV} (no pose in training and inference)  &  62.3 & 53.3 & 55.5 & 48.2 \\
        & \text{DP} + Ours (no pose in inference) &  66.0  & 58.7 & 56.9 & 51.2  \\
        & \text{DP} + Concat (with pose in inference) 
        & \textbf{67.5} & \textbf{60.2} & \textbf{58.1} & \textbf{55.1}\\
    \midrule
        \text{Desktop}\cite{desktop} 
        & \text{DP}\cite{Ghoddoosian_2023_ICCV} (no pose in training and inference)  &  10.5 & 5.1 & 36.8 & 2.3 \\
        & \text{DP} + Ours (no pose in inference) &  18.0 & 7.6 & \textbf{52.2} & 3.7  \\
        & \text{DP} + Concat (with pose in inference) & \textbf{19.1} & \textbf{9.3} & \textbf{52.2} & \textbf{8.3} \\
    \bottomrule
    \end{tabular}}
    \label{tab:fusion_strategy_table}
\end{table}

\begin{table}[t]
    \caption{Offline weakly-supervised segmentation results on the Desktop dataset with TASL as the baseline.}
    \centering
    \begin{tabular}{l|l|l|c|c|c}
    \toprule
         Training & Testing & \textit{acc} & \textit{IoU} & \textit{Edit} & \textit{F1@0.5} \\
    \midrule
        Pose & Pose & 28.7 & 18.2 & 91.6 & 11.3 \\
        RGB+Pose & Pose & 30.2 & 19.5 & 93.4 & 12.6 \\
        RGB & RGB & 35.2 & 22.4 & 95.8 & 14.7 \\
        RGB+Pose & RGB & 41.2 & 27.1 & 96.6 & 20.7 \\
        
    \bottomrule
    \end{tabular}
    \label{tab:only_pose_table}
\end{table}

\begin{figure*}[t]
\centering
 \includegraphics[width=0.95\textwidth,keepaspectratio]{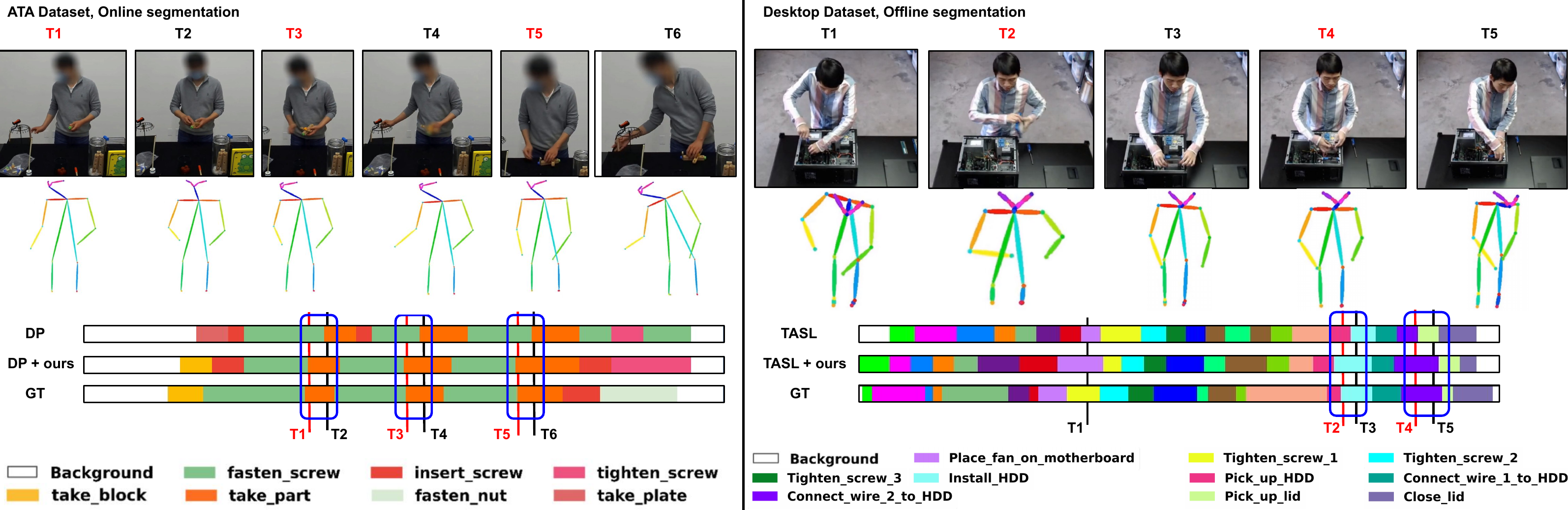}
\caption{Qualitative results of our pose-based contrastive learning in online (left) and offline (right) segmentation. Understanding fine-grained human pose results in more accurate detection of action boundaries at test time.}
\label{fig:quali}
\end{figure*}

\subsection{Qualitative Analysis}
Fig. \ref{fig:quali} shows that enabling the RGB encoder to understand human poses enhances the accuracy of segmentation models. This improvement is attributed to the model's ability to learn nuances of human poses and their variations within the same segment or during transitions from one action to another. Effectively, the detection of action boundaries becomes more precise. For example, in the ATA dataset, frames at time $T_1$ and $T_2$ show that the action ``take part'' consists of two main poses: extending a hand to grasp the part and placing it on the block. The baseline fails to identify the first pose as part of the action ``take part'', whereas our method has learned the extension pose precedes the placing pose as part of a single action. This pattern is consistently observed three times in Fig. \ref{fig:quali} (left). 

Additionally, note the  fine granularity of poses that the RGB encoder can capture in videos from Desktop Assembly. Particularly, our method is able to recognize the transition from ``pickup HDD'' at time $T_2$ to ``install HDD'' at time $T_3$. Also, our method is able to recognize that the combination of the two poses at time $T_4$ and time $T_5$ correspond to the action ``connect wire''. In this instance, the absence of pose knowledge in the baseline results in incorrect  detection of an action transition.  It is remarkable that that such detailed pose understanding is obtained without explicitly utilizing the pose modality during inference. Yet, there are still instances where the infused pose information is not sufficiently discriminative to accurately identify the correct action. For example, at time $T_1$  the action ``tighten screw'' is incorrectly classified as ``place fan''.
\section{Limitations}
\label{sec:limitations}
The proposed paper sheds light on the impact of off-the-shelf pose estimation in weakly-supervised action segmentation. However, it suffers from two main limitations. Firstly, the choice of $\delta$ is dependent on the viewpoint. Because the relative distance between joints vary across different viewpoints, finding an optimal value that works best from different viewpoints can be challenging. Secondly, our method is devised for single-person action segmentation, so in videos with background people, e.g. IKEA dataset, it requires additional heuristics to eliminate background poses.

\section{Conclusion}
\label{sec:conclusion}

We introduce a weakly supervised action segmentation framework that leverages human pose knowledge in long instructional videos. The framework explores interactions between video sequences and human pose sequences during training and avoids using pose features at  inference. Extensive experiments  demonstrate the efficacy of the method as it outperforms the previous SOTA in segmenting long instructional videos under both online and offline settings. Furthermore, our framework can be extended to various segmentation backbones, pose extractors, causal and non causal settings for several representative datasets.
{
    \small
    \bibliographystyle{ieeenat_fullname}
    \bibliography{main}
}

\clearpage
\setcounter{page}{1}
\maketitlesupplementary

\section{More Implementation Details}

\begin{table*}[tb]
    \caption{Split-wise comparison of proposed method versus baseline on IKEA dataset for online action segmentation.}
    \centering
    \resizebox{\textwidth}{!}{
    \begin{tabular}{l|c|c|c|c}
    \toprule
        Metric & acc & IoU & Edit & F1@0.5 \\
    \midrule
        Split & 1/2/3/4/5 & 1/2/3/4/5 & 1/2/3/4/5 & 1/2/3/4/5 \\
    \midrule
        \text{Greedy \cite{CDFL}} & 54.4/60.1/\textbf{50.9}/\textbf{54.9}/45.1 & 28.5/30.8/26.2/\textbf{29.6}/20.2 & \textbf{48.3}/46.7/37.4/42.2/33.0 & 22.7/28.1/21.8/\textbf{26.1}/19.7 \\ 
        \text{DP \cite{Ghoddoosian_2023_ICCV}} &  56.6/59.6/50.2/51.8/\textbf{53.1} & 28.3/30.7/\textbf{26.3}/26.2/\textbf{24.9} & 46.8/\textbf{55.3}/46.2/47.2/\textbf{45.0} & 24.9/29.9/\textbf{24.5}/25.0/\textbf{25.9} \\
        \text{DP + Ours}
        & \textbf{57.3}/\textbf{61.7}/50.3/51.3/51.4
        & \textbf{29.9}/\textbf{31.5}/\textbf{26.3}/26.6/24.2
        & \textbf{48.3}/\textbf{55.3}/\textbf{46.3}/\textbf{47.6}/44.6
        & \textbf{25.9}/\textbf{30.2}/24.2/25.2/25.5 \\ 
    \bottomrule
    \end{tabular}}
    \label{tab:ikea_dataset_comparison_online}
\end{table*}

\subsection{Implementation Details on Pose Normalization}
Given a frame at time $t$, raw pose $p_t \in \mathbb{Z}^{K\times 2}$ is a collection of $(x,y)$ coordinates for $K$ human keypoints. Here, $K$ represents the number of 2D keypoints extracted by an external pose extractor and $\mathbb{Z}$ is the set of integers. Before inputting these raw keypoints to the pose encoder, we perform a normalization step to ensure they are unaffected by changes in perspective, rotation, and positional offset in the frame. Specifically, each keypoint is centered and scaled with respect to the "center of mass" of the human, which is determined by averaging the coordinates of all joints. Subsequently, we determine the angle required to rotate each adjusted keypoint so that the head and "center of mass" align vertically, sharing the same $x$ coordinates. The specific mathematical formulation is listed below:

\begin{align*}
\text{centroid} = \frac{1}{K} \sum_{i=1}^{K} p_{t_i} \\
\text{centered\_pose} = p_t - \text{centroid} \\
\text{avg\_distance} = \frac{1}{K} \sum_{i=1}^{K} \sqrt{(x_{t_i} - x_{\text{centroid}})^2 + (y_{t_i} - y_{\text{centroid}})^2} \\
\text{scaled\_pose} = \frac{\text{centered\_pose}}{\text{avg\_distance}} \\
\text{angle} = \arctan\left(\frac{y_{\text{scaled\_pose\_of\_head\_joint}}}{x_{\text{scaled\_pose\_of\_head\_joint}}}\right) \\
\text{rotation\_matrix} = \begin{bmatrix} \cos(\text{angle}) & -\sin(\text{angle}) \\ \sin(\text{angle}) & \cos(\text{angle}) \end{bmatrix} \\
\text{normalized\_pose} = \text{scaled\_pose} \times \text{rotation\_matrix}
\end{align*}

These normalized 2D keypoints, $\overline{p}_t$, are then fed into the pose encoder. 

Our pose network is a fully-connected MLP network with sizes [34, 128, 128, output\_size], where the output\_size is determined by the specific network architecture we use in training. 

\subsection{Implementation Details on different backbones} 
As mentioned in the main paper, we extracted I3D \cite{Carreira2017QuoVA} features from ATA and IKEA datasets, and for Desktop Assembly, we used ResNet \cite{ResNet} features. The dimension for I3D features in ATA dataset is 2048, whereas in IKEA is 400. The dimension of ResNet feature in Desktop dataset is 512. 

In experiments with DP\cite{Ghoddoosian_2023_ICCV} as the baseline, we modeled the video encoder with Transformers. The projection network for video feature is a fully-connected layer of input size that is determined by the input dimension of video features and output size of 128. We set the pose network to have input size of 34 and output size of 128 to have a matched dimension for contrastive learning. During inference time, the projection and pose networks are not used. The detailed parameters of network structure are not changed. In our experimentation, the learning rate is set to 0.01, beam size is 151, window size is 15. During evaluation, we use the default exploration threshold of 0.7 for our segmentation results on ATA dataset. Also, we set an exploration threshold of 0.0 for IKEA and Desktop datasets due to their similar training and test transcripts. The training iteration is 40000 for ATA dataset, 20000 for IKEA dataset, and 10000 for Desktop dataset.

In experiments with TASL\cite{tasl}, we regard the existing GRU network as the output for RGB embedding. The output dimension of RGB embedding is 64, so we set the pose network to have input size of 34 and output size of 64 to perform contrastive learning. In our experimentation, we simply add the contrastive learning loss without any network modification. Specifically, in the TASL architecture, the learning rate is 0.01, decode sample rate is 30, window size is 33, space size is 10, pred size is 3, auto encoder weight is 0.2, edge window is 6 and edge step is set to 2. The training iteration is 20000 for ATA dataset and 6000 for Desktop Dataset.

For MuCon\cite{mucon2021}, we pass the scaled pose keypoints to the pose encoder to obtain pose embeddings of size 2048, corresponding to the RGB embeddings. These RGB embeddings are produced by a multi-stage temporal convolutional network \cite{mstcn}. However, we pass the pose embeddings to a ``frozen'' copy of the temporal convolutional network to obtain pose embeddings that correspond to the same format as the RGB embeddings, i.e., same number of embeddings in time and same dimensionality. Then, we perform the contrastive learning on these embeddings for both pose and RGB modalities. In our experiments, we train for 100 epochs for both the baseline and our method. The specific parameters are set to their default values with learning rate of 0.01, and momentum of 0.0. It is noteworthy to mention that MuCon has three output versions, and we picked the best version (MuCon-full) for our comparisons.

\section{Experimental Results on IKEA Dataset}
As mentioned in the main paper, we provide split-wise results in Table \ref{tab:ikea_dataset_comparison_online}. The overall results in the main paper are computed as the average of all splits. We associate the overall marginal improvements on the IKEA dataset mostly to its 5th split. For other splits, single-person is mostly exhibited in the training and testing sets. On the contrary, in many videos of the 5th split, the single-person assumption is violated by background people, which negatively impacts our pose encoding accuracy. While our contrastive learning module only establishes RGB-pose correspondence for each person, the pose encoding might not be so accurate when there are multiple persons in background. Results of split three and split four are competitive between our method and the baseline, whereas splits one and two exhibit the largest improvements of our proposed pose-infused methodology. In general, our method beats previous baselines in most cases in the IKEA dataset over different metrics and splits.

\end{document}